\documentclass{article}
\usepackage{graphicx}
\usepackage{setspace}
\usepackage{amsmath}
\usepackage{fixltx2e}
\usepackage{natbib,stfloats}
\usepackage{mathrsfs}
\usepackage{enumitem}
\usepackage{makecell}

\setcellgapes{4pt}

\begin{document}










\title{Maximizing Store Revenues using Tabu Search for Floor Space Optimization}




\author{Jiefeng Xu \and Evren Gul \and Alvin Lim}
\date{Research and Development Department\\ Precima, a Nielsen Company}

\maketitle

\begin{abstract}
Floor space optimization (FSO) is a critical revenue management problem commonly encountered by today's retailers. It maximizes store revenue by optimally allocating floor space to product
categories which are assigned to their most appropriate planograms. We formulate the problem as a connected
multi-choice knapsack problem with an additional global constraint and propose a tabu search based metaheuristic that exploits the
multiple special neighborhood structures. We also incorporate a mechanism to determine how to combine the multiple neighborhood moves. A
candidate list strategy based on learning from prior search history is also employed to improve the search quality. The results of
computational testing with a set of test problems show that our tabu search heuristic can solve all problems within a reasonable amount of time. Analyses of individual contributions of relevant components of the algorithm were conducted with computational experiments.
\end{abstract}

\section{Introduction}
Floor space is a valuable and scarce asset for retailers. Over the last decade, the number of products competing for limited space
increased by up to 30\% \citep{ehi1994retail}. Thus, the efficient allocation of store floor space to product categories to maximize the total
store revenue can provide a significant edge to retailers in an increasingly competitive industry. Consequently, floor space management is considered as one of the vital strategic levers for retail revenue management \citep{kimes2011role}.

\begin{figure}[htb]
  \centering
  \includegraphics[width=4.8in]{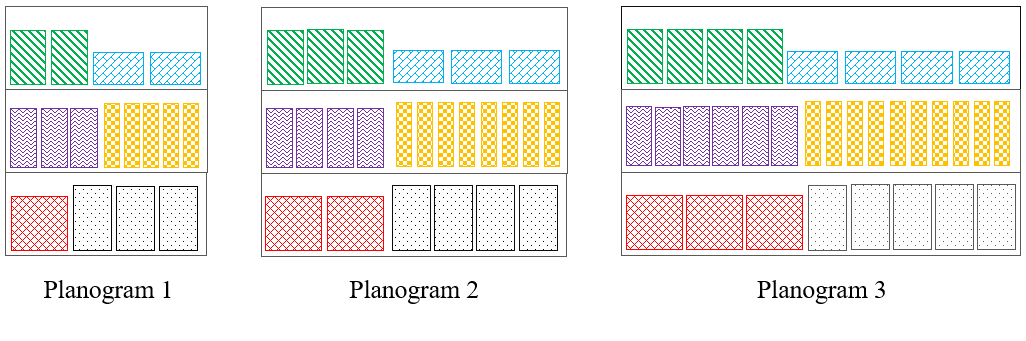}
  \caption{Sequence of planograms for a category with item facings and locations on shelves}
  \label{fig:plano}
\end{figure}

The problem we addressed is additionally motivated by the floor space
planning operations of a major grocery chain in Europe. In space planning operations, the grocer benefits from software (e.g. the JDA
Planogram Generator). Space planners of the grocer first create template planograms using software
to determine product placement on shelves. A planogram is a visual diagram that shows positions and the number of so-called facings of items
(corresponding to visible items). The product mix of categories, merchandising rules, sales patterns and characteristics of display furniture and
fixture are considered in the preparation of planograms. An active planogram in a store is replaced by an associated planogram
sequence. Figure \ref{fig:plano} shows a sequence of three planograms on which pattern and color highlighted rectangles represent the items, their facings and locations on the shelves for a specific category. A store using the set of planograms for the category
in Figure \ref{fig:plano} can only increase or decrease the space allocation with respect to the order of the planograms.

Next, planners select a planogram for each relevant category and use software to automatically generate store layout. In the final step, planograms are
physically replicated on store shelves. Using space planning software, retailers save a substantial amount of time on these space planning
operations which otherwise would be carried out manually. Although space planning software provides efficient tools for visualizing and
preparing of store layout and planograms and day-to-day maintenance and reporting activities, they are limited in incorporating
the effect of space on sales and neglect to fully take advantage of mathematical optimization in floor space decisions. Planners decide the assignment of  planograms to categories according to their
experience and by employing heuristics (based on criteria sometimes called ``proportional-to-market share'', and ``proportional-to-profit share'').  However, these
decisions can be far from optimal \citep{desmet1998estimation}. First of all, space elasticity, the ratio of change in sales to change
in space, is not considered. Secondly, the predicted revenue associated with various planograms is not known and not used to determine the best combination of planograms
in a store to maximize total revenue.

Therefore, in our work we created an integrated solution using advanced
mathematical modeling techniques in order to maximize revenue of each store. Our approach for FSO includes two steps: (1) develop a
statistical model to measure the space elasticity; and (2) formulate and solve an optimization
problem for each store to determine the optimal assignment of planograms to maximize total revenue subject to certain business constraints. Our statistical model is able to predict revenue of a product category for a given planogram, and in turn, the mathematical optimization
efficiently determines the best assignment of planograms to categories.  In this paper, our focus is on the design and implementation of a solution approach based on tabu search to solve the optimization problem in step (2).

In the early work on space modeling, the emphasis was placed on establishing a relation between space and sales. Indeed, the positive
impact of space allocation on sales has been documented by several studies \citep{curhan1972relationship, corstjens1981model, bultez1989asymmetric, borin1994model, dreze1994shelf, desmet1998estimation}.  In light of research on space modeling, we developed
a statistical model to measure the effect of allocated space in a planogram on category sales focusing on the solution of the space management optimization problem specified in Step (2) mentioned above. We refer the reader to \cite{hubner2012retail}, \cite{eisend2014shelf}, \cite{kok2015assortment} and \cite{bianchi2018allocating} for holistic reviews of approaches to modeling of space effects.

In the second step of FSO, we maximize store revenues by optimally assigning planograms in stores. Modern revenue management launched with the airline industry and it advanced with the applications in two other traditional industries: hotels and car rental companies \citep{chiang2007overview}. The success in these traditional industries attracted others and revenue management has since been applied in industries like restaurants, cargo, cruise, subscription services, theme parks and retail. More recently, revenue management has been introduced to newer industries such as cloud computing, home delivery, rideshare and e-commerce \citep{klein2020review}. Books by \cite{talluri2006theory} and \cite{phillips2005pricing}\ provide introduction to concepts and aspects of revenue management including pricing, capacity allocation, network management, overbooking and markdown management and review papers by \cite{chiang2007overview} and \cite{klein2020review} provide comprehensive surveys on the developments in revenue management over the last 40 years.

In the revenue management literature, FSO is considered as an application in the retail industry. Early applications of revenue management in retail started with seasonal items which are analogous to perishability of airline seat inventory. The value of the seasonal items diminish significantly after the selling season. Therefore, \cite{coulter2001decreasing} proposed the maximization of revenue by applying optimal discount pricing to seasonal items. In another study, \cite{aviv2005partially} investigated the dynamic pricing for fashion-like goods for a seasonal retailer. In their second work \cite{aviv2008optimal}, they extended their work to optimal pricing of seasonal goods in the presence of strategic customers. Practitioners \cite{hawtin2003practicalities} and \cite{lippman2003retail} provided overall guidelines on the implementation of revenue management systems in grocery retail outlets. They mainly focused on pricing strategy and discussed potential benefits and challenges in systems integration. In the recent years, revenue management for online retailers has been studied. \cite{agatz2013revenue} identified differentiation in price and delivery options as a key for revenue maximization. \cite{belavina2017online} analyzed the effect of subscription and per-order delivery revenue models on sales and environment for online grocery retailers.

As mentioned earlier, \cite{kimes2011role} emphasized that space is the third strategic lever in revenue management, along with the price and time levers. They pointed out that space management is less studied in the literature, but is equally important as management of price and time. The aforementioned literature for retail revenue management mainly focuses on ``price'' and ``time'' levers and only includes ``space'' implicitly in some of the studies, i.e., none of the research works explicitly studied the effects of space allocation in optimizing revenues. Even though retailers manage time and price well, the performance will be sub-optimal in the context of revenue management if they do not manage their space well \citep{kimes2011role}. In our problem, all the levers are taken into account in the space-effect and optimization model where we explicitly optimize space by maximizing revenue per linear meter of a grocery store space for the implementation time period. FSO is an essential and integral component of any revenue management system that will enable retailers to achieve their revenue potential.

In sum, the floor space optimization problem considered here involves the optimal allocation of the available planograms in a store to maximize total
predicted revenue. The problem is subject to constraints due to planogram sequencing, store layout, furniture and fixture
characteristics. The first constraint requires that, for each existing planogram in a store, a replacement planogram should be chosen from its planogram sequence.
This constraint ensures that every product category currently in the store is reassigned a planogram after
the optimization. Space planners group the planograms with respect to furniture and fixture requirements and location in the store.
These groups are called planogram worlds (PWs). The second constraint requires that the sum of lengths of planograms in each PW is bounded by lower and
upper total length limits. Finally, the third constraint arises from the fact that a store can accommodate space expansion up to a certain level and requires space above some threshold. Therefore, this constraint imposes lower and upper limits on the total length of all planograms in the store (Additional background on the connection of FSO to other developments is given in the Appendix).

Intuitively, the FSO problem seems similar to the maximization version of the well-known knapsack problem (KP), as well as its
variants such as multiple knapsack problems, or multiple choice knapsack problems, since we precalculate the expected revenues of planograms. If
we view a planogram as an item, and a planogram world as a knapsack or bin, the FSO problem clearly contains the resource
constraints (like the length limits) which are similar to  knapsack constraints. For a comprehensive review on KP and its
variants, we refer readers to \cite{hiremath2008new}. However, the additional global store length constraint of our FSO problem adds more
complexities to the already complicated KP.

This paper develops a tabu search metaheuristic algorithm that exploits the specific neighborhood structures of the FSO
problem. The next section defines the mathematical formulation of the FSO problem, and then provides a
relevant literature review. In Section~3, we describe the tabu search algorithm specifically. The computational results are included
in Section~4. Finally, we summarize our findings in Section~5.

\section{Mathematical Formulation for FSO}

The space model predicts revenue for a given space and other predictor variables of a product category in a planogram. Thus, we can
define the predicted revenue of product category $i$ when assigned planogram $j$, $R_{ij}$ as
\begin{align*}
R_{ij} = h(s_{ij},{\mathbf{u}}_{ij}),
\end{align*}
where $h(\cdot)$  represents the statistical model, $s_{ij}$ is the space assigned to category $i$ with planogram $j$, and $\mathbf{u}_{ij}$ is the vector of
values for other predictor variables. Note that for the purposes of this study, the $R_{ij}$ values are precomputed and therefore assumed to be constants.
The FSO problem can then be formulated as a mixed integer programming problem as follows.\\

\noindent
\textbf{Index:}
\begin{itemize}
  \item $I$: set of product categories
  \item $J$: set of planograms
  \item $K$: set of planogram worlds
  \item $J_i$: set of planograms that can be assigned to category $i$ where $\bigcap_{i \in I}J_i=\emptyset$
  \item $I_k$: set of categories belonging to PW $k$
  \item $J_k$: set of planograms belonging to PW $k$ where $J_k=\bigcup_{i \in I_{k}}J_i$
\end{itemize}

\noindent
\textbf{Constants:}
\begin{itemize}
  \item $R_{ij}$ : the revenue of category $i$ if assigned to planogram $j \in J_i$
  \item $L_{ij}$: the length (shelf space) for category $i$ if assigned to planogram $j \in J_i$
  \item $LL_k$: the lower bound of the total length for PW $k$
  \item $UL_k$: the upper bound of the total length for PW $k$
  \item $LS$: the lower bound on sum of all planogram lengths in the entire store
  \item $US$: the upper bound on sum of all planogram lengths in the entire store
\end{itemize}

\noindent
\textbf{Decision Variables:} $x_{ij}$: the binary variable with value 1 if category $i$ is assigned to planogram $j$, and 0 otherwise.\\

\noindent
\textbf{Model:}
\begin{alignat}{2}
  \textrm{Maximize} \quad &\sum_{i \in I} \sum_{ j \in J_i} R_{ij} x_{ij}&& \label{eqn:objective}\\
  \textrm{subject to} \quad &\sum_{j \in J_i} x_{i,j} =1, &\quad &\forall i \in I, \label{eqn:assign_planogram}\\
  &\sum_{i \in I_k} \sum_{j \in J_i} L_{ij} x_{ij} \geq LL_k,  &\quad & \forall k \in K, \label{eqn:pw_lower_limit}\\
  &\sum_{i \in I_k} \sum_{j \in J_i} L_{ij} x_{ij} \leq UL_k,  &\quad & \forall k \in K, \label{eqn:pw_upper_limit}\\
  &\sum_{i \in I} \sum_{j \in J_i} L_{ij} x_{ij} \geq LS, && \label{eqn:store_lower_limit}\\
  &\sum_{i \in I} \sum_{j \in J_i} L_{ij} x_{ij} \leq US, && \label{eqn:store_upper_limit}\\
  &x_{ij}  \in \{0,1\}, &\quad & \forall i \in I, j \in {J_i}. \label{eqn:binary_var}
\end{alignat}

The objective function (\ref{eqn:objective}) maximizes the total store revenue which is the sum
of revenues of all categories placed on planograms. The constraint (\ref{eqn:assign_planogram}) stipulates each category should be assigned to a planogram. The constraints (\ref{eqn:pw_lower_limit}) and
(\ref{eqn:pw_upper_limit}) establish the lower and upper length limits for each PW, and constraints (\ref{eqn:store_lower_limit}) and (\ref{eqn:store_upper_limit}) enforce the lower and upper limits of
total planogram length for the store. The constraint (\ref{eqn:binary_var}) defines the binary variable for $x_{ij}$. Without the presence of constraints
(\ref{eqn:store_lower_limit}) and (\ref{eqn:store_upper_limit}), the problem would be equivalent to solving $|K|$ multiple-choice KPs.

It is well known that the KP is NP-hard \citep{karp1972reducibility}. The literature on KP and its variants is rich. Exact methods have focused on employing
branch and bound and dynamic programming approaches \citep{martello1985algorithm, martello1997upper, martello2003exact, martello1999dynamic, pisinger1995minimal, pisinger1997minimal, pisinger1999core, pisinger1999exact}. A variety of heuristics and metaheuristics have been designed for solving practical problems quickly, including those based on
genetic algorithm \citep{chu1997genetic, raidl1998improved}, tabu search \cite{glover1996critical,lokketangen1998solving}, ant
colony algorithms \citep{shi2006solution}, simulated annealing \citep{liu2006improved}, global harmony search \citep{zou2011solving}, etc. In addition to the papers that have proposed algorithms, \cite{pisinger2005hard} conducted an interesting study on how to design test problems that appeared to be hard for several exact methods. Since FSO embeds a multiple choice KP-like
NP hard subproblem, it is natural to develop a metaheuristic-based approach such as the tabu search algorithm for solving this practical problem.

\section{The Tabu Search Algorithm for FSO}
The tabu search (TS) algorithm is a well known metaheuristic for solving a large number of both theoretical
and practical optimization problems. It employs adaptive memory to overcome the limitation of conventional search methods such as hill-climbing, which terminate (and hence
become “trapped”) in a locally optimal solution within the current neighborhood. The common mechanisms TS employs include short term memory, long term memory, aspiration rule, intensification and diversification strategies.  For a more comprehensive compendium of TS and its advanced strategies, we refer readers to \cite{glover1997tabu}.

Our TS algorithm for FSO (denoted as TSFSO) starts from an initial solution and evaluates the objective function by calculating the
revenues from planogram assignments and penalties from all length violations. Our method employs multiple simple neighborhood structures, and determines the moves based on these neighborhoods at each iteration through a scenario based control
mechanism (controller). To improve the efficiency of the TS and reduce the effort spent on examining inferior solutions, we devise a
learning-based candidate list strategy which benefits from the statistics collected from the search history.  These
components are elaborated in the next few subsections.

\subsection{Initial Solution and Objective Function Evaluation}
Our initial solution is constructed based on the following three simple rules:
\begin{enumerate}
\item Least length rule: each category is assigned to the planogram in which it occupies the least length. 


This rule ensures that violations of all upper limit length constraints will be minimized, but violations of the lower
limit length constraints may occur.

\item Highest revenue rule: each category is assigned to the planogram where it will yield the highest revenue. 


This rule will produce a solution that achieves an upper bound on the revenue that can be obtained by an optimal FSO solution. However, violations to length constraints may occur  coming from both upper and lower limits of the length constraints.

\item Balanced rule: each category is assigned to the planogram that yields  the maximal revenue per length unit. 

This rule will generate a more balanced solution by considering  both revenue and length requirements, though it may
still cause violations of length constraints.
\end{enumerate}

Let $G_k(x)$ denote the total length occupied by the current assignment $x$, that is $G_k(x)=\sum_{i \in I_k} \sum_{j \in J_i}
L_{ij} x_{ij}$. Then we combine the revenue and violation into a single objective function in the TSFSO as follows:
\begin{align*}
f(x)=&\sum_{i \in I} \sum_{ j \in J_i} R_{ij} x_{ij} \\
& -P\left(\sum_{k \in K}\left(\max(0,G_k(x) - UL_k)+\max(0,LL_k- G_k(x)) \right)\right.\\
& \left. + \max(0,\sum_{k \in K} G_k(x) - US) + \max(0, LS - \sum_{k \in K} G_k(x))\right)
\end{align*}
In the above objective, $P$ is a large positive number. When the assignment $x_{ij}$ is changed at each iteration, the
value of $f(x)$ is recalculated accordingly. Upon termination, the $x_{ij}$ that produced the maximum value $f(x)$ is considered as the
best solution. If the violation term associated with a solution is zero, then the solution is feasible.

\subsection{Neighborhood Moves}
\begin{figure}[htb]
  \centering
  \includegraphics[width=3.0in]{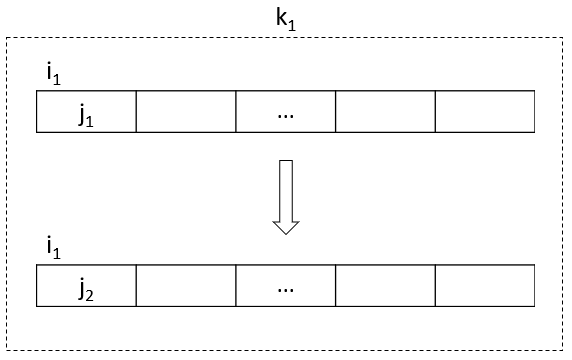}
  \caption{An Example of Level 1 Move}
  \label{fig:move1}
\end{figure}

The core decision of the FSO is to assign a planogram for each category to its corresponding PW. The neighborhood move is performed by assigning different planograms to categories iteratively. We design the five basic neighborhood moves as follows:

\noindent
\textbf{Level~1 Move:} Select a category and move it from its current planogram to another planogram. In PW $k_1$, let $i_1$ be the category under
consideration, let $j_1$ ($j_1 \in J_{i_1} $) be the planogram that $i_1$ is currently assigned to, and $j_2$ is the new planogram
($j_2\neq j_1$, $j_2 \in J_{i_1} $). Then the Level 1 Move changes the assignment $x_{i_1j_1}=1, x_{i_1j_2}=0$ to $x_{i_1j_1}=0, x_{i_1j_2}=1$.
Such a move results in the following changes in objective function evaluation:
\begin{align*}
\Delta f(x)=&R_{i_1j_2}-R_{i_1j_1}-P(\max(0,G_{k_1}(x)+L_{i_1j_2}-L_{i_1j_1}-UL_{k_1}) \\
&+ \max(0, LL_{k_1} -G_{k_1}(x)-L_{i_1j_2}+L_{i_1j_1}) + \max(0,\sum_{k \in K}G_k(x) \\
&+L_{i_1j_2}-L_{i_1j_1} -US)
 + \max(0,LS-\sum_{k \in K}G_k(x) -L_{i_1j_2}+L_{i_1j_1}))
\end{align*}
An example of a Level 1 Move is visually illustrated in Figure \ref{fig:move1}.

Once a Level 1 Move is performed, i.e., the category $i_1$ (in PW $k_1$) is moved from planogram $j_1$ to $j_2$, a tabu restriction is
applied to prevent $i_1$ from being moved back to $j_1$ within a specific number of iterations. The duration (in iterations) of such a restriction is called the $tabu$ $tenure$ of the move, and is customarily selected randomly between a lower and upper bound.  After a Level 1 Move (moving $i_1$ from $j_1$ to $j_2$), the reversed move (changing $x_{i_{1}j_1}=0$ to $x_{i_{1}j_1}=1$ is made $tabu$ for all iterations starting from the current iteration, $currIter$, through a last iteration $T(i,j)$ for $i=i_1$ and $j=j_1$ by the assignment:
\begin{equation*}
    T(i_1,j_1)=currIter + random(TL1,TH1)
\end{equation*}
\noindent
where the value $random(TL1,TH1)$ identifies the tabu tenure based on the lower and upper bounds, $TL1$ and $TH1$. The value $T(i_1,j_1)$ may be called the (short-term) $tabu$ $memory$ for the move.\\

\begin{figure}[h]
  \centering
  \includegraphics[width=3.0in]{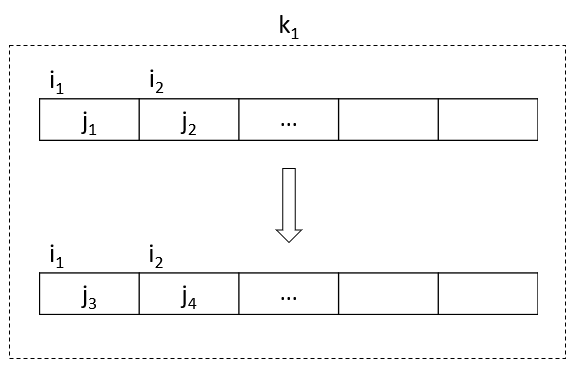}
  \caption{An Example of Level~2 Move}
  \label{fig:move2}
\end{figure}

\noindent
\textbf{Level~2 Move:} Select two categories from their current planograms (in  PW $k_1$) to different planograms in the same PW. Let
$i_1$, $i_2$ ($i_1\neq i_2$) be the two categories under consideration, let $j_1$, $j_2$ ($j_1 \in J_{i_1}, j_2 \in J_{i_2}$) be the two
currently assigned planograms for $i_1$ and $i_2$, and $j_3$ and $j_4$ be the new planograms ($j_3\neq j_1$, $j_3 \in J_{i_1}$ ,
$j_4\neq j_2$, $j_4 \in J_{i_2}$ ) for $i_1$ and $i_2$. Then the Level~2 Move changes the decision variable values from $x_{i_1j_1}=1, x_{i_2j_2}=1, x_{i_1j_3}=0, x_{i_2j_4}=0$ to
$x_{i_1j_1}=0, x_{i_2j_2}=0, x_{i_1j_3}=1, x_{i_2j_4}=1$. Such a move results in the following changes in objective function
evaluation:
\begin{align*}
\Delta f(x)=&R_{i_1j_3}+R_{i_2j_4}-R_{i_1j_1}-R_{i_2,j_2}-P(\max(0,G_{k_1}(x)+L_{i_1j_3}+L_{i_2j_4}\\
&
-L_{i_1j_1}-L_{i_2j_2}-UL_{k_1}) + \max(0, LL_{k_1} -G_{k_1}(x)-L_{i_1j_3}-L_{i_2j_4} \\ &+L_{i_1j_1}+L_{i_2j_2})
 + \max(0,\sum_{k \in K}G_{k}(x) +L_{i_1j_3}+L_{i_2j_4}-L_{i_1j_1}-L_{i_2j_2} \\
 &-US) + \max(0,LS-\sum_{k \in K} G_{k}(x) -L_{i_1j_3}-L_{i_2j_4}+L_{i_1j_1}+L_{i_2j_2}))
\end{align*}
Figure \ref{fig:move2} shows an example of a Level~2 Move.

The tabu memory for a Level~2 Move (that moves category $i_1$ from planogram $j_1$ to $j_3$, and category $i_2$ from planogram $j_2$ to $j_4$)
is to enforce $x_{i_1j_1}=0$ for all iterations  less than or equal to $ T(i_1,j_1)$ and $x_{i_2j_2}=0$ for all iterations less than or equal to $ T(i_2,j_2)$. After such a
Level~2 Move, the tabu memory is updated as:
\begin{align*}
T(i_1,j_1)&= currIter + random(TL2,TH2)\\
T(i_2,j_2)&= currIter + random(TL2,TH2)
\end{align*}

\begin{figure}[h]
  \centering
  \includegraphics[width=3.0in]{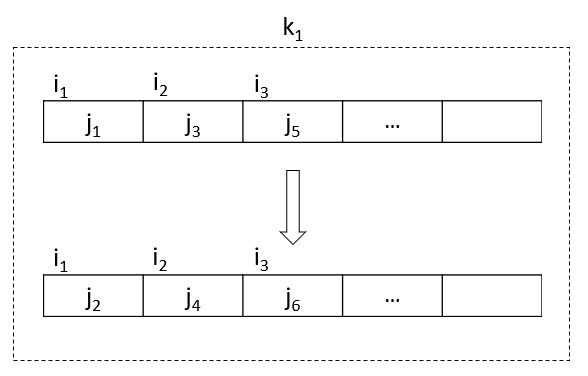}
  \caption{An Example of Level~3 Move}
  \label{fig:move3}
\end{figure}

\noindent
\textbf{Level~3 Move:} Reassign three categories from their current planograms (in PW $k_1$) to different planograms in the same PW. Let
$i_1$, $i_2$ and $i_3$ ($i_1\neq i_2$, $i_2\neq i_3$, $i_3 \neq i_1$ be the three categories under consideration, let $j_1$, $j_2$ and $j_3$
($j_1 \in J_{i_1}$, $j_2 \in J_{i_2}$, $j_3 \in J_{i_3}$) be the three currently assigned planograms for $i_1$, $i_2$ and $i_3$,
$j_4$, $j_5$ and $j_6$ be the new planograms ($j_4 \neq j_1$, $j_5 \neq j_2$, $j_6 \neq j_3$, $j_4 \in J_{i_1}$, $j_5 \in J_{i_2} $
, $j_6 \in J_{i_3}$) for these categories. Then the Level~3 Move changes the decision variable values from $x_{i_1j_1}=1$, $x_{i_2j_2}=1$, $x_{i_3j_3}=1$, $x_{i_1j_4}=0$, $x_{i_2j_5}=0$,
$x_{i_3j_6}=0$ to $x_{i_1j_1}=0$, $x_{i_2j_2}=0$, $x_{i_3j_3}=0$, $x_{i_1j_4}=1$, $x_{i_2j_5}=1$, $x_{i_3j_6}=1$. Such a move
results in the following changes in the objective function evaluation:
\begin{align*}
\Delta f(x)=&R_{i_1j_4}+R_{i_2j_5}+R_{i_3j_6}-R_{i_1j_1}-R_{i_2j_2}-R_{i_3j_3}-P
(\max(0,G_{k_1}(x)\\
&+L_{i_1j_4}+L_{i_2j_5}
+L_{i_3j_6}-L_{i_1j_1}-L_{i_2j_2}-L_{i_3j_3}-UL_{k_1}) \\
&+ \max(0, LL_{k_1}
-G_{k_1}(x)-L_{i_1j_4}-L_{i_2j_5}-L_{i_3j_6}+L_{i_1j_1}+L_{i_2j_2}
\\
& +L_{i_3j_3}) + \max(0,\sum_{k \in K}G_k(x) + L_{i_1j_4}+L_{i_2j_5}+L_{i_3j_6}-L_{i_1j_1}-L_{i_2j_2}\\
&-L_{i_3j_3} -US) + \max(0,LS-\sum_{k \in K}G_{k}(x) -L_{i_1j_4}-L_{i_2j_5}-L_{i_3j_6}\\
&+L_{i_1j_1}+L_{i_2j_2} +L_{i_3j_3}))
\end{align*}
Figure \ref{fig:move3} displays an example of a Level~3 Move.

Similarly, after a Level~3 Move (that moves the category $i_1$ from planogram $j_1$ to $j_4$, and category $i_2$ from planogram $j_2$ to $j_5$,
and category $j_3$ from planogram $j_3$ to $j_6$) , the tabu memory is updated by setting:
\begin{align*}
T(i_1,j_1)&=currIter + random(TL3,TH3)\\
T(i_2,j_2)&=currIter + random(TL3,TH3)\\
T(i_3,j_3)&=currIter + random(TL3,TH3)
\end{align*}

\begin{figure}[h]
  \centering
  \includegraphics[width=4.8
  in]{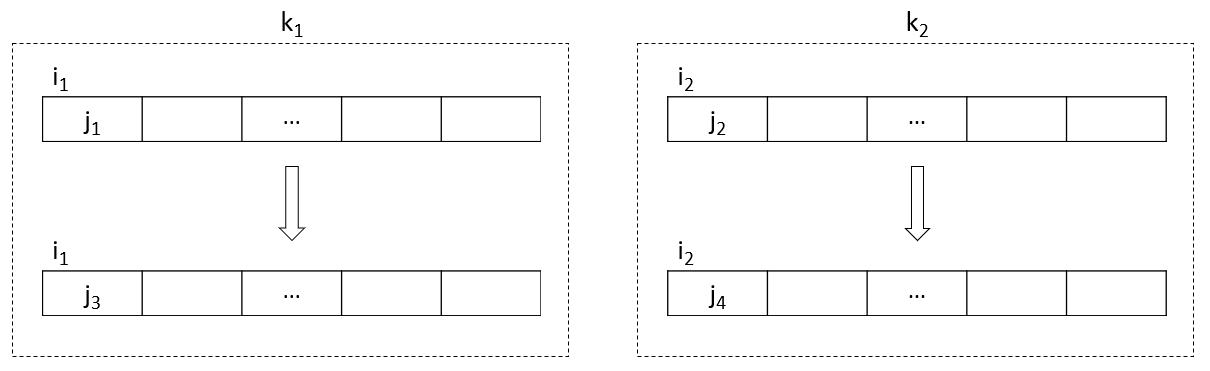}
  \caption{An Example of Level~4 Move}
  \label{fig:move4}
\end{figure}

\noindent
\textbf{Level~4 Move:} Construct a composite move that performs the two Level~1 Moves simultaneously for two PWs. Let $i_1$ (in PW
$k_1$) and $i_2$ (in PW $k_2$) be the two categories under consideration, let $j_1$ ($j_1 \in J_{i_1} $) and $j_2$ ($j_2 \in J_{i_2} $) be the
planograms that hold $i_1$ and $i_2$, respectively. Finally, let $j_3$ and $j_4$ be the new planograms for $i_1$ and $i_2$ ($j_3\neq j_1$,
$j_3 \in J_{i_1} $, $j_4\neq j_2$, $j_4 \in J_{i_2}$) for $i_1$ and $i_2$. Then the Level~4 Move changes the decision variable values from $x_{i_1j_1}=1$, $x_{i_1j_3}=0$,
$x_{i_2j_2}=1$, $x_{i_2j_4}=0$ to $x_{i_1j_1}=0$, $x_{i_1j_3}=1$, $x_{i_2j_2}=0$, $x_{i_2j_4}=1$. Such a move results in the
following changes in the objective function evaluation:
\begin{align*}
\Delta f(x)=&R_{i_1j_3}-R_{i_1j_1}+ R_{i_2j_4}-R_{i_2j_2} -P(\max(0,G_{k_1}(x)+L_{i_1j_3}\\
&-L_{i_1j_1}-UL_{k_1})+\max(0,G_{k_2}(x)+L_{i_2j_4}-L_{i_2j_2}-UL_{k_2})\\
&+ \max(0, LL_{k_1} -G_{k_1}(x)-L_{i_1j_3}+L_{i_1j_1})+ \max(0, LL_{k_2} \\
&-G_{k_2}(x)-L_{i_2j_4}+L_{i_2j_2})+ \max(0,\sum_{k \in K}G_k(x) +L_{i_1j_3}\\
&-L_{i_1j_1} +L_{i_2j_4}-L_{i_2j_2} -US) + \max(0,LS-\sum_{k \in K}G_k(x) \\
&-L_{i_1j_3}+L_{i_1j_1} -L_{i_2j_2}+L_{i_2j_4}))
\end{align*}
Figure \ref{fig:move4} depicts an example of a Level~4 Move where categories $i_1$ and $i_2$ are assigned to new planograms in the two PWs.

The tabu memory for a Level~4 Move (that moves category $i_1$ from planogram $j_1$ to $j_3$, and category $i_2$ from planogram $j_2$ to $j_4$)
is updated as follows.
\begin{align*}
T(i_1,j_1)&=currIter + random(TL4,TH4)\\
T(i_2,j_2)&=currIter + random(TL4,TH4)
\end{align*}

The difference between this and a Level 2 Move is that in the latter, the two categories $i_1$ and $i_2$ are located in the same PW, while in a Level~4 Move, they are located in the different PWs.\\

\begin{figure}[h]
  \centering
  \includegraphics[width=4.8in]{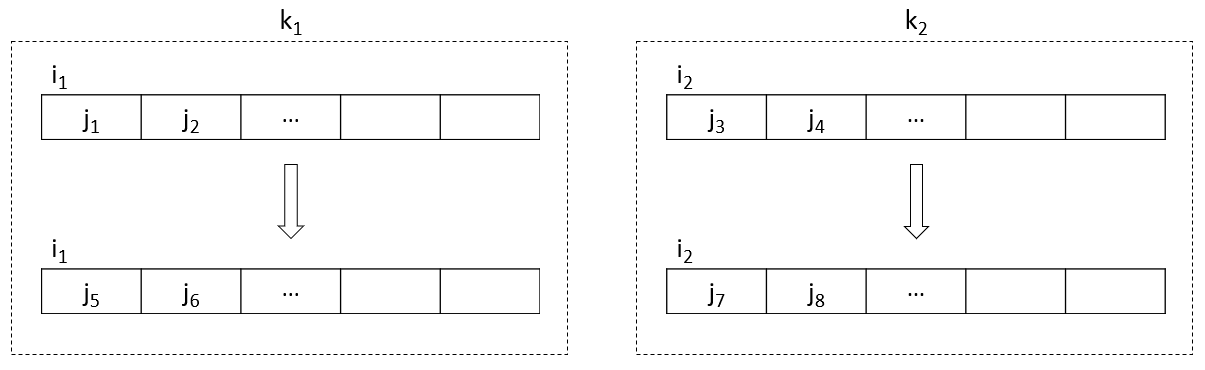}
  \caption{An Example of Level~5 Move}
  \label{fig:move5}
\end{figure}

\noindent
\textbf{Level~5 Move:} Like the Level~4 move, this move involves two PWs, in each PW two categories change their current planograms to the new planograms.
Let $i_1$, $i_2$ (in PW $k_1$) and $i_3$, $i_4$ (in PW $k_2$) be the categories under consideration, and let $j_1$ ($j_1 \in J_{i_1} $), $j_2$
($j_2 \in J_{i_2} $), $j_3$ ($j_3 \in J_{i_3} $) and $j_4$ ($j_4 \in J_{i_4} $) be the planograms that hold $i_1, i_2, i_3$ and
$i_4$, respectively. Finally, let $j_5$ and $j_6$ be the new planograms for $i_1$ and $i_2$ ($j_5\neq j_1$, $j_5 \in J_{i_1} $, $j_6\neq
j_2$, $j_6 \in J_{i_2}$), and let $j_7$ and $j_8$ be the new planograms for $i_3$ and $i_4$ ($j_7\neq j_3$, $j_7 \in J_{i_3}$,$j_8\neq
j_4$, $j_8 \in J_{i_4}$ ). Then the Level~5 Move changes the decision variable values from $x_{i_1j_1}=1$, $x_{i_1j_5}=0$, $x_{i_2j_2}=1$, $x_{i_2j_6}=0$,
$x_{i_3j_3}=1$, $x_{i_3j_7}=0$, $x_{i_4j_4}=1$, $x_{i_4j_8}=0$ to $x_{i_1j_1}=0$, $x_{i_1j_5}=1$, $x_{i_2j_2}=0$, $x_{i_2j_6}=1$,
$x_{i_3j_3}=0$, $x_{i_3j_7}=1$, $x_{i_4j_4}=0$, $x_{i_4j_8}=1$. The changes in the objective function value are identified by :
\begin{align*}
\Delta f(x)=&R_{i_1j_5}-R_{i_1j_1}+ R_{i_2j_6}-R_{i_2j_2} + R_{i_3j_7}-R_{i_3j_3}+ R_{i_4j_8}-R_{i_4j_4} \\
&-P(\max(0,G_{k_1}(x)+L_{i_1j_5}-L_{i_1j_1}+L_{i_2j_6}-L_{i_2j_2}-UL_{k_1}) \\
& + \max(0,G_{k_2}(x)+L_{i_3j_7}-L_{i_3j_3}+L_{i_4j_8}-L_{i_4j_4}-UL_{k_2})\\
&+ \max(0, LL_{k_1} -G_{k_1}(x)-L_{i_1j_5}+L_{i_1j_1}  -L_{i_2j_6}+L_{i_2j_2})\\
&+ \max(0, LL_{k_2}-G_{k_2}(x)-L_{i_3j_7}+L_{i_3j_3}-L_{i_4j_8}+L_{i_4j_4})\\
& +\max(0,\sum_{k \in K}G_k(x) +L_{i_1j_5}-L_{i_1j_1} +L_{i_2j_6}-L_{i_2j_2}+ L_{i_3j_7}\\
&-L_{i_3j_3} + L_{i_4j_8}-L_{i_4j_4}-US)+ \max(0,LS-\sum_{k \in K}G_k(x) \\
&-L_{i_1j_5}+L_{i_1j_1}-L_{i_2j_6}+L_{i_2j_2}-L_{i_3j_7}-L_{i_3j_3}\\
&-L_{i_4j_8}-L_{i_4j_4})) \\
\end{align*}
Figure \ref{fig:move5} exhibits an example of a Level~5 Move where categories $(i_1,i_2)$ and $(i_3,i_4)$ are assigned to new planograms in the
two PWs.

Similarly, the tabu memory for a Level~5 Move (that moves category $i_1$ from planogram $j_1$ to $j_5$, category $i_2$ from planogram $j_2$ to
$j_6$, category $i_3$ from planogram $j_3$ to $j_7$, and category $i_4$ from planogram $j_4$ to $j_8$) is updated as:
\begin{align*}
T(i_1,j_1)&=currIter + random(TL5,TH5)\\
T(i_2,j_2)&=currIter + random(TL5,TH5)\\
T(i_3,j_3)&=currIter + random(TL5,TH5)\\
T(i_4,j_4)&=currIter + random(TL5,TH5)
\end{align*}

Note that in TSFSO, once a type of neighborhood move is determined at each iteration, all legal moves from its respective neighborhood space
are evaluated, and the move with the best evaluation subject to the corresponding tabu restrictions. If a better solution than the
best solution found so far is detected, an $aspiration$ $rule$ permits the move to be performed regardless of the tabu memory restrictions. Other types of aspiration rules are also sometimes used, but the preceding rule is most often favored.

The type of move selected in TSFSO is determined by a scenario based control mechanism described next.

\subsection{The Scenario-based Controller}
Our use of multiple neighborhood moves, where each neighborhood provides a special structure for moving from one solution to another, is a type of strategy that is a critical element in meta-heuristics such as variable neighborhood search (VNS) \citep[see][]{mladenovic1997variable}. However, multiple neighborhoods strategies had already been successfully implemented in TS applications before the first VNS paper was published \citep[see][]{glover1984heuristic, xu1996using, xu1996network}. In contrast to the VNS mechanism that allows the search to iterate over different neighborhoods, in such TS applications, a special control mechanism is designed to determine when to apply a specific neighborhood most efficiently and effectively. An example of this relevant to our current algorithm also appears in \cite{xu1998fine}.

In our five level neighborhoods, the lower level neighborhood moves are more effective for intensification to improve the current solution in regions around local optima, while the higher level neighborhood moves cause greater structural changes that move farther from a current solution and induce diversification. On the other hand, the number of move alternatives associated with the lower level neighborhoods is  smaller than the number of alternatives associated with  the higher level neighborhoods, and  so the evaluation of the former is faster than that of the latter.  Based on these observations, we design the following rules for the scenario-based controller by considering the balance between intensification and diversification, together with the trade-offs for efficiency. In the rules described below, a lower numbered rule can be overwritten by a higher numbered rule if, as indicated, this is valid for a specific scenario:
\begin{description}[labelindent=\parindent,leftmargin=\parindent]
  \item [\textbf{Rule~1:}] During the first stage of the search ($T_1$ iterations), the Level~2 neighborhood is used to quickly improve the solution.
  \item [\textbf{Rule~2:}] During the second stage of the search ($T_2$ iterations), the lower level neighborhood moves are used probabilistically based on the probabilities as follows: $p_1$ for Level~1 neighborhood move, $p_2$ for Level~2 neighborhood move, and $p_3$ for Level~3 neighborhood move.
  \item [\textbf{Rule~3:}] During second stage of the search, whenever a new best ever solution is found, we ``downgrade'' neighborhood move type for the next iteration to reduce the current neighborhood move by one level. (This forces the search to focus more fully on  intensification in an attempt to find an even better solution than the current new best solution.)
  \item [\textbf{Rule~4:}] At any iteration, if the current best solution could not be improved after a certain number of iterations denoted by $T_3$ (which often signals a search deadlock that requires a diversification strategy to break the impasse), a high level neighborhood (Level~4 or 5) move is required. The selection of the neighborhood is based on the probabilities $p_4$ for a Level~4 neighborhood and $p_5$ for a Level~5 neighborhood.
  \item [\textbf{Rule~5:}] If level~4 or 5 is selected per Rule~4, then this level cannot be performed for more than $T_4$ consecutive iterations. Once $T_4$ consecutive iterations of Level~4 or 5 moves are used, then the search must be switched back to a lower level neighborhood (Level~1,2,or 3, the exact type to use is determined probabilistically by Rule~2.) for at least $T_5$ iterations. (Rule~5 prevents the search from focusing too long on diversification strategies when these strategies are not successful in improving the best solution. This allows the search to strategically oscillate between the intensification and diversification strategies.)
  \item [\textbf{Rule~6:}] The search terminates either when the maximum number of iterations is reached ($T_0=T_1+T_2$), or when the search cannot improve the current best feasible solution for $T_6$ iterations.
\end{description}

In  Rules~2 and 4, a randomization based probabilistic selection method is used for iterating between the different types of
neighborhoods. This allows the search to overcome the strong local optimality tendency produced by adhering to a single neighborhood and
amplifies the diversification effect. The scenario-triggered rules (which are invoked when (1) a
quick improvement is required during early search; (2) a new best solution is found; (3) the search fails to find a new best solution for a certain
time, etc.) are applied to enhance the ability to exploit the special
neighborhood structures of various neighborhoods, while accounting for issues of effectiveness and efficiency.

\subsection{The Learning-based Candidate List Strategy}
To further improve the efficiency of TSFSO, we consider the fact that a good neighborhood move may contain attribute values similar to those of
good neighborhood moves performed in the past. For this, we employ a $candidate$ $list$ to guide our choice of moves based on keeping track of attributes contained in
good past moves.

We collect statistics from the moves performed at each of the first $T_7$ iterations in the search history and count the frequency of
the PW/category combinations of the moves executed at each iteration. Let $U_{ki}$ be the number of times that PW $k$ and category $i$ are
involved in the moves performed so far, where initially $U_{ki}$ is set to zero, and $U_{ki}$ is updated at each iteration as
follows.

\begin{table}[htb]
\centering \makegapedcells
\caption{Move Attributes Statistics Collection}
\label{table:moves}
\begin{tabular}{|l|c|l|}
  \hline
  \thead{Performed \\ Move Type} & \thead{Involved \\ PW/Category} & \thead{Update $U_{ki}$} \\
  \hline
  Level~1& $(k,i)$ & $U_{ki}=U_{ki}+1$\\
  \hline
  Level~2& $\begin{array} {l} (k,i_1)\\ (k,i_2) \end{array} $ &  $\begin{array}{l} U_{ki_1}=U_{ki_1}+1\\U_{ki_2}=U_{ki_2}+1
  \end{array} $  \\
  \hline
  Level~3& $\begin{array} {l} (k,i_1)\\ (k,i_2)\\(k,i_3) \end{array} $ &  $\begin{array}{l}
  U_{ki_1}=U_{ki_1}+1\\U_{ki_2}=U_{ki_2}+1\\U_{ki_3}=U_{ki_3}+1 \end{array} $  \\
\hline
   Level~4& $\begin{array} {l} (k_1,i_1)\\ (k_2,i_2) \end{array} $ &  $\begin{array}{l}
   U_{k_1i_1}=U_{k_1i_1}+1\\U_{k_2i_2}=U_{k_2i_2}+1 \end{array} $  \\
\hline
   Level~5& $\begin{array} {l} (k_1,i_1)\\ (k_1,i_2)\\(k_2,i_3)\\(k_2,i_4) \end{array} $ &  $\begin{array}{l}
   U_{k_1i_1}=U_{k_1i_1}+1\\U_{k_1i_2}=U_{k_1i_2}+1\\U_{k_2i_3}=U_{k_2i_3}+1\\U_{k_2i_4}=U_{k_2i_4}+1 \end{array} $  \\
  \hline
\end{tabular}
\end{table}

After the statistics $U_{ki}$ are collected from the first $T_7$ iterations, we construct a candidate list at each iteration thereafter for
each type of neighborhood move performed at that iteration. The updating of the $U_{ki}$ values is continued, however,  until the end of
the search. For categories $I_k$ of PW $k$, let $n_k$ be the number of categories in $I_k$ that are included in the moves performed so far ($n_k
\leq |I_k|$), and let $\bar{U}_k=\{i'_1,i'_2,...,i'_{n_k}\}$ be a sorted list of category indices such that $U_{ki'_1,} \geq
U_{ki'_2} \geq ... \geq U_{ki'_{n_k}} >0$.

We further introduce a list $\bar{U}_{k}(n_k)$ containing the first $n_k$ category indices from $\bar{U}_k$ ($p \leq n_k$) (with $n_k$ identifying the
highest $U_{ki}$ value category). The candidate list construction is described in Table~\ref{table:candidate_list}.

\begin{table}[htb]
\centering  \makegapedcells
\caption{Candidate List Construction}
\label{table:candidate_list}
\resizebox{\textwidth}{!}{\begin{tabular}{|l|c|l|}
  \hline
  \thead{Performed \\ Move Type} & \thead{Involved \\ PW/Category} &
  \thead{Candidate List} \\
  \hline
  Level~1& $(k,i)$ & $\begin{array}{l}\mbox{No candidate list; use the full neighborhood,}\\ \mbox{i.e., } i \mbox{ is selected from
  } I_k.\end{array}$\\
  \hline
  Level~2& $\begin{array} {l}
  (k,i_1)
  \\ (k,i_2) \end{array} $ &
    $\begin{array}{l} \mbox{For PW } k, \mbox{if } |\bar{U}_{k}|>0 , \mbox{then } i_1, i_2 \mbox{ are selected from the}\\ \mbox{category
    list } \bar{U}_k(n_k*r_1);\\
  \mbox{Otherwise,} i_1, i_2 \mbox{ are selected from } I_k.
  \end{array}$   \\
  \hline
  Level~3& $\begin{array} {l} (k,i_1)\\ (k,i_2)\\(k,i_3) \end{array} $ &
  $ \begin{array}{l} \mbox{For PW } k, \mbox{if } |\bar{U}_{k}|>0 , \mbox{then } i_1, i_2, i_3 \mbox{ are selected from }\\
  \mbox{the category list } \bar{U}_k(n_k*r_1);\\
   \mbox{Otherwise, } i_1, i_2, i_3 \mbox{ are selected from } I_k. \end{array}$   \\
\hline
   Level~4& $\begin{array} {l} (k_1,i_1)\\ (k_2,i_2) \end{array}$  &
   $\begin{array}{l} \mbox{For PW } k_1, \mbox{if } |\bar{U}_{k_1}|>0, \mbox{then } i_1 \mbox{ is selected}\\
   \mbox{from the category list } \bar{U}_{k_1}(n_{k_1}*r_2);\\
   \mbox{Otherwise, } i_1 \mbox{ is selected from } IA_{s_1};\\
   \mbox{For PW } k_2, \mbox{if } |\bar{U}_{k_2}|>0, \mbox{ then } i_2 \mbox{ is selected from }\\
   \mbox{the category list } \bar{U}_{k_2}(n_{k_2}*r_2);\\
   \mbox{Otherwise, } i_2 \mbox{ is selected from } I_{k_2}.
     \end{array}$   \\
\hline
   Level~5& $\begin{array} {l} (k_1,i_1)\\ (k_1,i_2)\\(k_2,i_3)\\(k_2,i_4) \end{array}$  &  $\begin{array}{l}
    \mbox{For PW } k_1, \mbox{if } |\bar{U}_{k_1}|>0, \mbox{then } i_1, i_2 \mbox{ are selected from }\\
     \mbox{the category list } \bar{U}_{k_1}(n_{k_1}*r_2);\\
    \mbox{Otherwise, } i_1,i_2 \mbox{ are selected from } I_{k_1};\\
   \mbox{For PW } k_2, \mbox{if } |\bar{U}_{k_2}|>0 , \mbox{then } i_3,i_4 \mbox{ are selected from }\\
    \mbox{the category list } \bar{U}_{k_2}(n_{k_2}*r_2);\\
   \mbox{Otherwise, } i_3,i_4 \mbox{ are selected from } I_{k_2}
   \end{array}$   \\
  \hline
\end{tabular}}
\end{table}

In TSFSO, the ratios $r_1$ and $r_2$ indicate the degree of reduction from the full neighborhoods. The smaller the values for
$r_1$ and $r_2$, the more compressed is the neighborhood used for the respective candidate list.

\section{Computational Results}

TSFSO algorithm is implemented for the real floor space planning problem arising in a leading grocery chain in Europe. To investigate the effectiveness of TSFSO, we need to execute our TSFSO on a series of test problems. Since there are no known open benchmark problems available for FSO, we first design a method to generate a series of test problems.  We utilize store space configuration data from our real world implementation, as well as ideas for designing hard test problem instances for relevant knapsack problems in literature. We describe the test problem generation method in the next subsection.

\subsection{Test Problem Generation}

We choose a common store in the grocery chain with 9 PWs and 194 planograms and use its actual PW/planogram configuration as a basis for our test problem generation. To protect confidential information such as planogram length and revenue data for the company, we generated associated length ($L_{ij}$) and revenue ($R_{ij}$) data using an approach which aims to create difficult test instances for knapsack problems based on existing research in literature. More specifically, we use a method to create the so-called weakly correlated spanner instances with span(2,10) described in \cite{pisinger2005hard} to generate a series of coordinated pairs of length and revenue data as follows. We first generated a basic spanner set $(l_{\kappa}, r_{\kappa})$ for $\kappa\in \{1,2\}$ by setting $l_{\kappa}=random(1,10^7)$, and  $r_{\kappa}=random(l_{\kappa}-10^6,l_{\kappa}+10^6)$ for $\kappa \in \{1,2\}$. If the resulting $r_{\kappa} < 1$, we regenerate $r_{\kappa}$ until it satisfies $r_{\kappa} \geq 1$. Then we normalize the spanner set by setting $l_{\kappa}=\lceil l_{\kappa}/5\rceil, r_{\kappa}=\lceil r_{\kappa}/5\rceil$ for $\kappa \in \{1,2\}$. Then each pair of length and revenue numbers $(L_{ij},R_{ij})$ is generated by repeatedly taking one pair ($l_{\kappa},r_{\kappa}$) from the spanner set, and multiplying it by a value randomly generated from $[1,10]$.

In particular, we initialize the counter $k=0$ and for each valid combination of $(i,j)$, we identify the  spanner set index by $\kappa=k \textrm{ mod } 2 + 1$, $\alpha=random(1,10)$. The length and revenue numbers are then calculated as $L_{i,j}=\alpha l_{k}$ and $R_{ij}=\alpha r_{k}$, followed by incrementing the counter $k$ by setting $k=k+1$. We repeat the process until all possible length and revenue values $(L_{ij}, R_{ij})$ are generated for the required 194 planograms.

Unlike the method suggested in \cite{pisinger2005hard} where the length bounds are generated based on the generated lengths using different ratios across test instances (e.g., bounds are increased by a fixed percentage for each instance), however, we generated the lower and upper bounds for PWs and store levels for our test problems by multiplying fixed ratios (0.85 for lower bound and 1.15 for upper bound) with total (generated) lengths calculated based on the current planogram assignment. This approach may reduce the difficulty of the test problems but ensures the generated problems are feasible and bounds are consistent with the current store FSO planning practice as well. Nevertheless, the test problems we generated are still computationally more difficult than the real problem we encountered in practice, indicating that they are good problems to stress-test our TSFSO heuristic and to have some confidence that our heuristic can cover realistic problems we have yet to encounter.

We repeat the above process using different random seeds until we generate 100 test problems. To execute our TSFSO on these test problems, we use the parameter values described in the next subsection.

\subsection{Parameter Setting}
We set the values for the parameters based on a priori knowledge, as well as on the results from
limited brute-force experiments. Applying a systematic parameter fine-tuning method based on statistical analysis and experiment
design techniques may significantly improve the heuristic performance (see Xu $et$ $al.$ 1998b for an example).

First, the penalty $P$ for length violations (at both the PW level and store level) is set to $-20000$. In this application, the penalty
function is implemented as a static value,  heavily emphasizing feasibility over solution quality by strongly favoring feasible neighborhood moves, and hence focusing more intensification than diversification. We plan in a future work to design a dynamic, self-adaptive penalty parameter, to better explore the interplay between
intensification and diversification.

Tabu tenure parameters are set as follows: we unify such parameters for various neighborhood move types by designating a single
lower bound and a single upper bound. Such bounds are related to the respective neighborhood spaces so they can be resiliently
applied to small, medium or large neighborhoods. Specifically, for a given PW $k$, we set the lower bound value
$TL=TL1=TL1=TL3=TL4=TL5=\max\{4,|I_k|/2\}$, and the upper bound value $TH=TH1=TH1=TH3=TH4=TH5=TL+\min\{7,|I_k|/7\}$.

Several parameters govern the controller and the search progress. The entire search consists of 1200 iterations ($T_0=1200$), while the first stage of the search starts from iteration 1 to iteration $T_1=120$ and the second stage then invokes for the reset of $T_2=1080$ iterations. The parameter $T_3$, which triggers the condition for using high level neighborhood moves, is set to 20 iterations. The high level neighborhood moves can be performed no more than 2 consecutive iterations ($T_4=2$) and the next 10 iterations ($T_5=10$) will be dedicated to low level
neighborhood moves after 2 consecutive iterations for high level neighborhood moves. The TSFSO terminates either when the iteration
counter reaches $T_0=1200$, or when the current (feasible) best solution cannot be improved within the most recent $T_6=0.8*T_0$ iterations.

The probabilities used for selecting the different types of neighborhood moves are: $p_1=0.2, p_2=0.5, p_3=0.3, p_4=0.6,
p_5=0.4$. Note that $p_1+p_2+p_3=1.0$ and $p_4+p_5=1.0$.

Lastly, the candidate list strategy collects statistics from attributes of moves performed during the first 100 ($T_7=100$) iterations. Consequently, the candidate list strategy is initiated at the $101^{st}$ iteration. The two ratios used to reduce the neighborhood
spaces, are set to $r_1=0.5$, $r_2=0.8$, respectively.

\subsection{Computational Experiments and Results Analysis}
The TSFSO is implemented using the Python language and is executed on a Linux machine on cloud. The CPU model of the machine
is Intel\textsuperscript{\textregistered} Xeon\textsuperscript{\textregistered} CPU E5-2686 v4 @ 2.30GHz. To compare the effectiveness of TSFSO, we first use
Gurobi, one of the leading commercial solver packages for Mixed Integer Programming, to solve the 100 tests problems to optimality using the same computational environment. We limit the computation time to 200 seconds for each problem. Gurobi either finds an optimal solution before reaching the time limit or terminates with a best solution obtained at the time limit, which we designate it as the $de facto$ optimal solution, instead of classifying it as an optimal solution.

To evaluate the different time effect of TSFSO, we set the maximum number of iterations ($T_0$ in the four test runs of TSFSO to be 300, 600, 900, and 1,200. The associated test runs are accordingly denoted TSFSO-300, TSFSO-600, TSFSO-900, TSFSO-1200. All other parameters remain the same as previously described with the following features implemented:
\begin{itemize}
  \item used the balanced rule as the initial solution rule;
  \item used all 5 level moves coordinated by the scenario-based controller; and
  \item used candidate list.
\end{itemize}
In summarizing the results from across the 100 problems, we report the number of problems for which we obtain an optimal or $de facto$ optimal solution (as OPTNUM), the average percentage of the optimality gap across all 100 problems (as AVGGAP), maximum percentage of the optimality gap (as MAXGAP), and the total CPU time in seconds (as CPUTM) in Table~\ref{table:test100}.

\begin{table}[htb]
\centering \makegapedcells
\caption{Computational Results on 100 Test Problems}
\label{table:test100}
\begin{tabular}{|l|r|r|r|r|}
\hline
\thead{RUN} & \thead{OPTNUM} & \thead{AVGGAP(\%)} & \thead{MAXGAP(\%)}  & \thead{CPUTM} \\
\hline
TSFSO-300&35&0.23&0.96&59\\
TSFSO-600&68&0.08&0.52&154\\
TSFSO-900&83&0.04&0.52&217\\
TSFSO-1200&100&0.00&0.00&312\\
Gurobi&100&0.00&0.00&12,345\\
\hline
\end{tabular}
\end{table}

Table~\ref{table:test100} clearly demonstrates with the reported reasonable computational times that our TSFSO algorithm can overcome the computational intractability of the test problems and obtain exceedingly high quality solutions using just a fraction of the time required by the Gurobi solver. Although for iteration limits below 1,200, not all optimal solutions were attained, the maximum percentage gap is quite small at less than $1\%$, and these solutions were obtained by our TSFSO at significantly less computational time versus Gurobi. This  further confirms that TSFSO can be used as an effective practical optimization method for solving floor space optimization problems without relying on commercial proprietary solvers such as Gurobi. It should be noted that our TSFSO was implemented in Python, which is generally considered slower in computational speed as compared to the C language that Gurobi was implemented on.

Next, we run experiments on our TSFSO to determine the effects of the three different initial solution rules described in Section~3.1. We denote the TSFSO using the least length rule, the highest revenue rule, and the balanced rule as TSFSO-LL, TSFSO-HR, and TSFSO-1200, respectively. All such TSFSO variants use 1200 iterations which offers sufficient time to improve the solution quality using tabu search.

\begin{table}[htb]
\centering \makegapedcells
\caption{Comparison of Initial Solution Rules}
\label{table:testInit}
\begin{tabular}{|l|r|r|r|r|}
\hline
\thead{RUN} & \thead{OPTNUM} & \thead{AVGGAP(\%)} & \thead{MAXGAP(\%)}  & \thead{CPUTM} \\
\hline
TSFSO-LL&39&0.22&0.91&310\\
TSFSO-HR& 26 &0.33&1.40&314\\
TSFSO-1200&100&0.00&0.00&312\\
\hline
\end{tabular}
\end{table}

The results in Table~\ref{table:testInit} clearly shows that TSFSO with different initial solution rules can find very high quality solutions for FSO. Compared to the optimal (or $de facto$ optimal) solutions, the TSFSO using the balanced rules for generating initial solutions yields all optimal solutions for all cases, and the one using the least
length rule for initial solutions obtains 39 optimal solutions out of 100 cases. The TSFSO using the highest revenue rule based
initial solutions lags behind by obtaining 26 optimal solutions; however, the worst solution is only $1.4\%$ away from optimality (on average it is 0.33\%), indicating it can produce reliable high quality solutions for practical applications.

By carefully monitoring and comparing the search progress, we cannot conclude that the computation time is significantly sensitive to the choice of initial solution method. No matter which initial solution method is adapter, the TSFSO beats Gurobi by an obvious edge in computation time.

We use TSFSO-1200 as a basis for examining the effects of using multiple neighborhoods in TSFSO by comparing to the tests from those using a single type of neighborhood for Level~1, Level~2, Level~3, Level~4 and Level~5 moves, denoting these tests by TSFSO-1, TSFSO-2, TSFSO-3, TSFSO-4 and TSFSO-5 in Table~\ref{table:testType}. We also design a new test, designated TSFSO-NC, that deactivates the learning-based candidate list in TSFSO-1200. In addition to the values OPTNUM, AVGGAP, MAXGAP and CPUTM reported in the previous table, we also show in the column INFNUM the number of cases in which the runs fail to find a feasible solution when the algorithm terminated.

\begin{table}[htb]
\centering \makegapedcells
\caption{Comparison of Neighborhoods and Candidate List Strategy}
\label{table:testType}
\resizebox{\textwidth}{!}{\begin{tabular}{|l|r|r|r|r|r|}
\hline
\thead{RUN} & \thead{OPTNUM} & \thead{INFNUM} & \thead{AVGGAP(\%)$^*$} & \thead{MAXGAP(\%)$^*$}  & \thead{CPUTM} \\
\hline
TSFSO-1200&100&0&0.00&0.00&312\\
TSFSO-1&0&46&3.25&19.20&4\\
TSFSO-2&0&100&N/A&N/A&109\\
TSFSO-3&0&100&N/A&N/A&743\\
TSFSO-4&0&25&1.81&19.40&99\\
TSFSO-5&0&100&N/A&N/A&8,272\\
TSFSO-NC&91&0&0.02&0.48&2,731\\
\hline
\end{tabular}}
\raggedleft
$^*$ applies to feasible cases only
\end{table}

It is obvious in Table~\ref{table:testType} that the TSFSO version that uses multiple neighborhoods performs better than the versions equipped with a single type of neighborhood. None of the single neighborhood versions can compete with TSFSO-1200 in terms of the number of optimal/feasible solution obtained. As shown, the Level~1 neighborhood move is the fastest but is able to obtain feasible solutions only for 54 cases. Moreover, it yields  not a single optimal solution, and the average optimality gap of $3.25\%$. Level~4 neighborhood moves, which perform two Level~1 moves simultaneously, provide significant improvements in term of the feasible solutions obtained. The Level~2, 3 and 5 neighborhood moves, each of which changes more than one assignment for one or two PWs, can hardly find feasible solutions by solely relying on their own effort. This confirms that each type of move has its own unique merits and disadvantages for handling special problem structures. In combination, they complement each other to find superior solutions, as shown in our baseline TSFSO-1200 version.

Surprisingly, the high level Level~4 neighborhood moves perform slightly faster than the lower level Level~2 moves. We attribute this mainly to the fact that the Level~4 moves enable the algorithm to reach a feasible local optimal solution quickly in 75 cases. In each iteration, it may move to the first improving best solution without examining the whole neighborhood space. In contrast,  the Level~2 move version struggles to achieve feasibility, so it consequently searches the entire neighborhood space at each iteration, and finally terminates without finding a feasible solution.

The findings from the comparisons between TSFSO-1200 and TSFSO-NC also confirm that the learning-based candidate list strategy can effectively improve efficiency using only $11.4\%$ of the computation time required by the version without learning-based candidate list strategy (312 seconds versus 2,731 seconds). It is also interesting to note that the full
neighborhood search performed in TSFSO-NC does not always yield optimal solutions, though in the 9 cases it obtained solutions with exceedingly small optimality gaps.

Next, we examine the effects of tabu memory in TSFSO. We again use TSFSO-1200 as the base case and obtain varying cases to experiment as
follows: (a) no tabu memory is applied; (b) all tabu tenures are set to 4; (c) all tabu tenures are set to 7; (d)
$TL=TH=(\max\{4,|I_k|/2\}+min\{7,|I_k|/7\})/2$. The corresponding new tests are denoted TSFSO-a, TSFSO-b, TSFSO-c and TSFSO-d. Table~\ref{table:testMem} contains the results from these tests along with the baseline run TSFSO-1200.

\begin{table}[htb]
\centering \makegapedcells
\caption{Comparison of Different Tabu Memories}
\label{table:testMem}
\begin{tabular}{|l|r|r|r|r|r|}
\hline
\thead{RUN} & \thead{OPTNUM} & \thead{AVGGAP(\%)} & \thead{MAXGAP(\%)}  & \thead{CPUTM} \\
\hline
TSFSO-1200&100&0&0&312\\
TSFSO-a&0&0.87&2.5&458\\
TSFSO-b&16&0.39 &1.5&377\\
TSFSO-c&32&0.25&1.05&339\\
TSFSO-d&47&0.0.21&1.93&324\\
\hline
\end{tabular}
\end{table}

As shown in Table~\ref{table:testMem}, all tests obtain feasible solutions with marginal optimality gaps. This can be partially
attributed to the efficiency of the multiple neighborhood search method designed for TSFSO. The contribution of the tabu
memory to this outcome is quite noticeable. However, when the tabu memory is removed (TSFSO-a), the algorithm fails to obtain any optimal solutions in any of the 100 cases.
Upon implementing various tabu memory methods (TSFSO-b, TSFSO-c, TSFSO-d), the number of optimal solutions improved and the optimality gaps diminished for non-optimal cases. This
demonstrates the value of tabu memory for improving an already good multiple neighborhood procedure by providing the ability to escape from strong locally optimal solutions. The baseline version of our algorithm (TSFSO-1200) which determines the tabu tenure as indicated in Section 4.2 clearly outperforms the others.

In addition, based on the results shown in  Table~\ref{table:testMem}, tabu memory not only plays critical role in obtaining optimal or near optimal solutions, but also helpful in improving the efficiency with reduced computational time. This finding is consistent with our observation that those TSFSO variants without tabu memory or with simple implementation of tabu memory intend to use more complicated and time consuming neighborhood move types (i.e., type 3 or type 5 moves), since they may encounter search impasse more frequently. An appropriate design of tabu memory is one of the key factors for practical FSO applications that requires obtaining high quality solutions within reasonable amount of computing time.

\begin{figure}[h]
  \centering
  \resizebox{\textwidth}{!}{\includegraphics[width=5in]{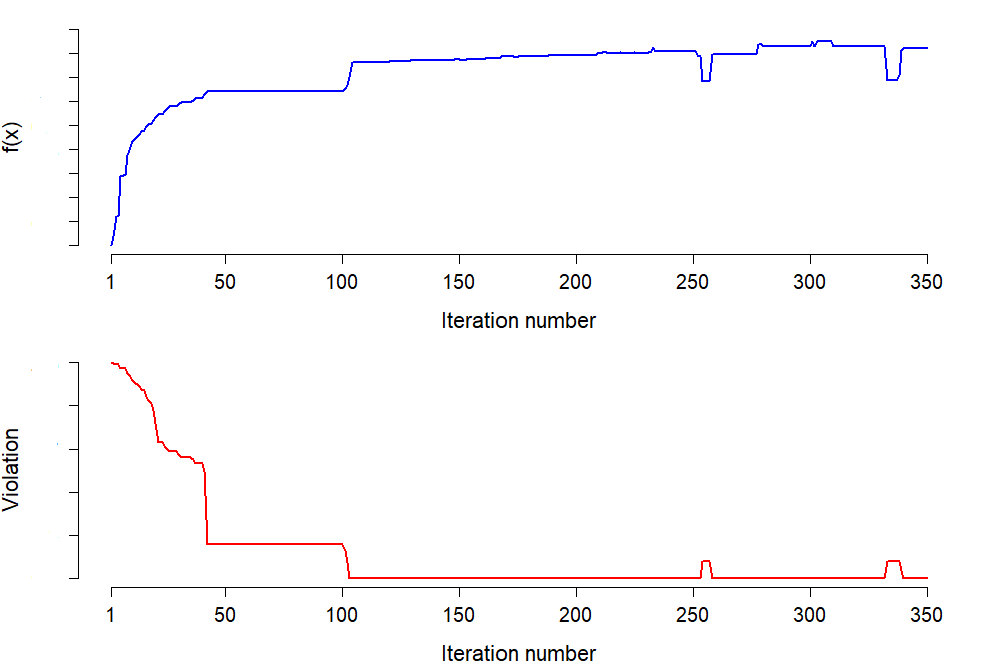}}
  \caption{A Search Progress Example }
  \label{fig:search}
\end{figure}

Finally, we demonstrate an example of the search progression of TSFSO for a selected run from our practical implementation in Figure~\ref{fig:search}. In this figure, the top and bottom charts show the value of the objective function $f(x)$ as well as the sum of the length constraint violations of the move performed at each iteration. (A zero violation indicates a feasible solution). The horizontal axis represents the iteration number, and the values of the vertical axis ($f(x)$ and violations) are transformed and re-scaled artificially (without loss of trends) for better illustrative purposes. Note the maximum number of iterations $T_0$ is set to 350 in this run.

Figure~\ref{fig:search} demonstrates that the TSFSO can rapidly improve solution quality after starting from an initial solution with a large feasibility violation. With the help of the large penalty, the solution becomes feasible at $103^{rd}$ iteration, and continues to improve via the multiple neighborhood moves. A locally optimal solution is obtained at the $233^{rd}$ iteration. Then the search continues, escaping the local optimum with the assistance of tabu memory. The search falls into an infeasible region at the $254^{th}$ iteration, and moves back into a feasible region at the $258^{th}$ iteration. The algorithm finds two new best solutions at the $278^{th}$ and $301^{st}$ iterations, and then again enters an infeasible region between iterations $333$ and $340$. No new best solutions were found throughout the remainder of the search, which terminated at the $350^{th}$ iteration.

\section{Conclusion}
We studied the floor space optimization problem arising in retail store revenue management by providing a mathematical formulation, conducting a review of relevant research and proposing a new solution approach based on tabu search. Our tabu search algorithm contains several innovative components, including a scenario-based controller for combining the multiple neighborhoods, and a candidate list strategy that utilizes learning based on statistics collected from prior search history.

We applied our TSFSO on 100 test problems that combined the practical configuration from real world applications as well as results from computational studies in literature. We reported results demonstrating that our method is highly effective and efficient in handling the test problems. Analysis of the experiment results further confirms the value of our scenario-based controller and our learning-based candidate list strategy. The use of the TSFSO in a practical application to a grocery chain improved its predicted annual revenue by around 1\%, which amounts to approximately 80 million Euros.

Future research work will continue in two directions: model improvement and algorithm improvement. First, we plan to enrich the space-effect and optimization models by considering planogram orientations, relative positions and layout. \cite{mowrey2018model, mowrey2019impact} studied the impact of layout and orientation of racks in a store on customer's navigation and exposure which provide guidance on how to allocate planograms in a store to maximize revenue. Because space is relatively fixed and scarce in a store, the inclusion of these additional aspects will improve the forecasting from the space-effect model and provide additional depth to the optimization model. Secondly, we will further improve the efficacy of TSFSO. Among the improvements envisioned, we plan to introduce a self-adaptive penalty function to replace the current reliance on a fixed penalty value that penalizes feasibility violations.  We also plan to study other potential improvements such as the analysis of the patterns of moves that prove most effective in order to further improve the search efficiency, the investigation of more advanced mechanisms to control the interplay between intensification and diversification, and the exploration of path relinking strategies to construct and exploit search trajectories between good solutions.

\section*{Acknowledgements}
The authors thank Professor Fred Glover for reviewing an early draft of this paper and providing helpful comments. The authors also thank the editor and two anonymous reviewers for their detailed and helpful comments.

\bibliographystyle{agsm}
\bibliography{TSFSO}

\end{document}